%% file: conference_041818.tex
\documentclass[letterpaper, 10 pt, conference]{ieeeconf}  

\IEEEoverridecommandlockouts                              
\input{tex/preamble}%
\usepackage{graphics}%
\usepackage{tikz,pgfplots}%
\usepackage{subfigure}%
\usetikzlibrary {arrows.meta} 
\newcommand{\AxisRotator}[1][rotate=0]{%
    \tikz [x=0.25cm,y=0.60cm,line width=.2ex,-stealth,#1,ultra thick] \draw (0,0) arc (-150:150:1 and 1);%
}
\usepackage{pgfplots}%
\usetikzlibrary{spy}

\title{\LARGE \bf
A Probabilistic Model of Activity Recognition with Loose Clothing
}

\author{Tianchen Shen$^{1}$, Irene Di Giulio$^{2}$ and Matthew Howard$^{1}$
\thanks{This work was supported by King's College London, the China Scholarship Council and \EPSRC     (EP/M507222/1). For the purpose of open access, the authors have applied a Creative Commons Attribution (CC BY) license to any Accepted Manuscript version arising.}
\thanks{$^{1}$Tianchen Shen ({\tt\small tianchen.shen@kcl.ac.uk}) and Matthew Howard are with \CORE, \DOE, \KCL, UK.}%
\thanks{$^{2}$Irene Di Giulio is with Centre for Human and Applied Physiological Sciences, \KCL, UK.}%
}

\begin{document}%
\makefinal
\maketitle%
\thispagestyle{empty}
\pagestyle{empty}
\begin{abstract}%
\input{abstract}\end{abstract}%

\section{Introduction}%
\input{introduction}%

\section{Probabilistic Modelling Framework}\label{s:problem_definition}%
\input{problem_definition}
\section{\Acrlong{ar} via Statistical Methods}\label{s:evaluation}%
\input{material_and_methods}
\section{Discussion}%
\input{discussion}%


{\scriptsize\printbibliography}%
\InputIfFileExists{sandpit.tex}{}{}
\end{document}

%% file: tex/preamble.tex
\usepackage{graphicx}\graphicspath{{figure/}}%
\usepackage{epstopdf}%
\usepackage{overpic}%
\usepackage[list=off]{caption}%
\usepackage{fmtcount}%
\usepackage{filecontents}%
\input{tex/abbreviations}%
\input{tex/symbols}%
\input{tex/format}%
\input{tex/comments}%
\makeatletter\newcommand{\manuallabel}[2]{\def\@currentlabel{#2}\label{#1}}\makeatother
\usepackage[%
backend=bibtex,%
style=ieee,%
url=false,%
doi=false,%
isbn=false,%
doi=true,%
eprint=false,%
]{biblatex}%

%% file: tex/abbreviations.tex
\input{ext/tex/abbreviations}

\newacronym{pd}{PD} {Parkinson's Disease}
\newacronym{mi}{MI} {Mutual information}
\newacronym{svm}{SVM}{support vector machine}
\newacronym{pcc}{PCC}{Pearson’s correlation coefficient}
\newacronym{ar}{AR}{activity recogntion}
\newacronym{me}{ME}{motion estimation}
\newacronym{gmm}{GMM}{Gaussian mixture model}
\newacronym{em}{EM}{Expectation maximization}
\newacronym{pdf}{PDF}{probability density function}
\newacronym{js}{JS}{Jensen-Shannon}
\newacronym{kl}{KL}{Kullback-Leibler}
\newacronym{cdf}{CDF}{cumulative distribution function}%
\newacronym{ks}{KS}{Kolmogorov-Smirnov}
\newacronym{mse}{MSE}{mean squared error}
\newacronym{etextiles}{e-textiles}{electronic textiles}
\newacronym{rmse}{RMSE}{Root mean squared error}

%% file: ext/tex/abbreviations.tex
\usepackage{xspace}
\usepackage[acronym]{glossaries}%
\glsdisablehyper 
\makeglossaries%

\newglossaryentry{rbf} 
{
	name={RBF},
	description={radial basis function},
	first={\glsentrydesc{rbf} (\glsentrytext{rbf})},
	plural={\glsentrytext{rbf}s},
	descriptionplural={\glsentrydesc{rbf}s},
	firstplural={\glsentrydescplural{rbf} (\glsentryplural{rbf})}
} 

\providecommand{\KCL    }{King's College London}

\providecommand{\CORE   }{the Centre for Robotics Research}

\providecommand{\DOE   }{Department of Engineering}

\providecommand{\EPSRC  }{Engineering and Physical Sciences Research Council}



%% file: tex/symbols.tex
\input{ext/tex/symbols}

\renewcommand  {\c}         {c}               
\providecommand{\y}         {c}               

\renewcommand  {\v}         {z} 
\providecommand {\w}        {\omega}

\providecommand {\xr}        {u}
\providecommand {\Xr}        {\MakeUppercase{\xr}}
\providecommand {\bxr}       {\mathbf{\xr}}
\providecommand {\xf}        {y}
\providecommand {\bxf}       {\mathbf{\xf}}
\providecommand {\Xf}        {\MakeUppercase{\xf}}
\providecommand {\df}        {F}
\providecommand {\dr}        {R}
\providecommand {\Df}        {\boldsymbol{\mathcal{\MakeUppercase{\xf}}}}
\providecommand {\Dr}        {\boldsymbol{\mathcal{\MakeUppercase{\xr}}}}
\providecommand {\DfAR}      {\Df}
\providecommand {\bxfd}      {\bxf^*} 
\providecommand {\yd}        {\y^*}  
\providecommand {\pf}        {p_\xf} 
\providecommand {\pr}        {p_\xr} 
\providecommand {\Xd}        {\Delta}
\providecommand {\xd}        {\delta}
\providecommand {\bdelta}    {\boldsymbol{\delta}}
\providecommand {\Dd}        {D}
\renewcommand   {\L}         {L}
\providecommand{\XfyO}       {\Xf_{\y=0}}
\providecommand{\Xfyl}       {\Xf_{\y=1}}
\providecommand{\XryO}       {\Xr_{\y=0}}
\providecommand{\Xryl}       {\Xr_{\y=1}}
\providecommand{\F}          {F} 
\providecommand{\FXfyO}      {\F(\XfyO)}
\providecommand{\FXfyl}      {\F(\Xfyl)}
\providecommand{\FXryO}      {\F(\XryO)}
\providecommand{\FXryl}      {\F(\Xryl)}
\providecommand {\fk}        {f_s} 
\providecommand {\Tk}        {T} 
\providecommand {\Nk}        {K} 
\providecommand{\dv}{\v_{21}}%
\providecommand{\vp}{\v_+}%
\providecommand{\vm}{\v_-}%
\providecommand{\Ay}{a}
\providecommand{\phiy}{\psi}
\providecommand{\Nw}{i}

\FPset {\fscotchYokeSimulation}{40}
\providecommand{\TkscotchYokeSimulation}{2\pi}
\FPset {\NdscotchYokeSimulationLo}{200}%
\FPset {\NdscotchYokeSimulationHi}{200}%
\FPeval{\NdscotchYokeSimulation}   {clip(\NdscotchYokeSimulationLo+\NdscotchYokeSimulationHi)}%
\FPset {\LminscotchYokeSimulation}{0}%
\FPset {\LmaxscotchYokeSimulation}{1}%
\FPset {\AscotchYokeSimulation}{1}%
\FPset {\kscotchYokeSimulation}{1}%
\FPset {\TrialsscotchYokeSimulation}{100}%
\FPset {\NdscotchYokeExperiment}{30}%
\FPset {\TkscotchYokeExperiment}{8}%
\FPset {\TrialsscotchYokeExperiment}{\TrialsscotchYokeSimulation}%
\FPset {\NwfkScotchYokeExperiment}{0.025}%
\FPset {\NdhandExperiment}{30}%

%% file: ext/tex/symbols.tex
\usepackage{amssymb}
\usepackage{mathtools}
\usepackage{amsmath}
\usepackage{gensymb}
\usepackage{amsfonts}

\mathchardef\mhyphen="2D   

\providecommand{\T}     {\top}                
\providecommand{\U}     {\mathcal{U}}         
\providecommand{\B}     {\mathcal{B}}         

\providecommand{\nd}      {n}                              
\providecommand{\Nd}      {\mathcal{\MakeUppercase{\nd}}}  
\providecommand{\nk}      {k}                              
\providecommand{\Nk}      {\mathcal{\MakeUppercase{\nk}}}  

\providecommand{\x}      {x}                  

\providecommand{\u}      {u}                  

\providecommand{\nu}     {q}                  

\providecommand{\bphi}  {\boldsymbol{\phi}}               
\providecommand{\bPhi}  {\boldsymbol{\Phi}}               %



%% file: tex/format.tex
\input{ext/tex/format}

\usepackage{xcolor}
\colorlet{xr}{blue}
\colorlet{xf}{red}

%% file: ext/tex/format.tex
\usepackage{xspace}
\usepackage{siunitx} 
\usepackage{nameref}

\providecommand{\eg}{\textit{e.g.,}~} %
\providecommand{\ie}{\textit{i.e.,}~} %
\providecommand{\qv}{\textit{q.v.}} %
\providecommand{\etc}{\textit{etc.}} %
\providecommand*{\sref}[1]{\S\ref{s:#1}}            
\providecommand{\tablename}{Table}
\providecommand{\figurename}{Fig.}
\providecommand*{\fref}[1]{\figurename~\ref{f:#1}}  
\providecommand*{\eref}[1]{(\ref{e:#1})}            

\let\labelindent\relax 
\usepackage[inline]{enumitem} 
\setlist{nolistsep}
\providecommand{\il}[1]{\begin{enumerate*}[label=(\roman*)]#1\end{enumerate*}} 
\providecommand{\cl}[1]{\begin{enumerate*}[label=(\alph*)]#1\end{enumerate*}}  

%% file: tex/comments.tex
\input{ext/tex/comments}
\usepackage{soul}

\colorlet{id}{blue}
\colorlet{ts}{green}

\renewcommand{\commentcolourcode}{Comments colour code: 
 {\color{mh}MH}%
,{\color{id}ID}%
,{\color{ts}TS}%
} 


%% file: ext/tex/comments.tex
\usepackage[normalem]{ulem}                                        
\usepackage{marginnote}
\usepackage[textwidth=\marginparwidth,colorinlistoftodos]{todonotes}
\providecommand{\tinytodo}[2][]
{\todo[caption={#2}, size=\small, #1]{\renewcommand{\baselinestretch}{0.6}\selectfont#2\par}}

\colorlet{jb}{red}
\colorlet{mh}{red}
\providecommand{\commentcolourcode}{Comments colour code: 
{\color{mh}MH}} 
\providecommand  {\colorsout}[1]{\bgroup\markoverwith{\textcolor{#1}{\rule[0.5ex]{2pt}{0.4pt}}}\ULon} 

\providecommand{\atJB} {{\color{jb}@JB}}

\providecommand  {\done}   [1]{\sout{#1}}
\providecommand  {\edit}   [3]{\colorsout{#3}{#1}{\color{#3}{#2}}}
\providecommand  {\nomment}[2]{\tinytodo[color=white,nolist,linecolor=#2,bordercolor=white,noinline]{\protect{\scriptsize\color{#2}#1}}}
\providecommand  {\note}[2]{\tinytodo[color=white,linecolor=#2,bordercolor=white,noinline]{\protect{\scriptsize\color{#2}#1}}}

\providecommand{\makecomments}{%
\section*{Usage notes}~\\
\noindent Use \texttt{\textbackslash note\{your comment\}\{your initials\}} to add comments/to dos. For example,
\underline{J}oe \underline{B}loggs adds comments using \texttt{\textbackslash note\{Comment.\}\{jb\}}.\nomment{These comments will appear as margin notes. They will also appear in the todo list on the first page.}{jb} \\[2ex]
You can mark a comment as resolved using \texttt{\textbackslash done\{Comment text.\}}, (\eg \texttt{\textbackslash note\{\textbackslash done\{\textbackslash atJB: Please fix this.\}\}\{mh\}}). It will then be formatted like this: \done{\atJB: Please fix this.}.\\[2ex]
\commentcolourcode\\[1ex]
\printacronyms~\\[2ex]
\listoftodos~\\[2ex]
\clearpage\setcounter{page}{1}
}

\providecommand{\makenotes}{\clearpage\appendix\input{notes}}%

\providecommand{\makefinal}{%
\renewcommand{\makecomments}{}\renewcommand{\makenotes}{}\renewcommand{\note}[1]{}\renewcommand{\done}[1]{}\renewcommand{\edit}[3]{\#2}%
}

%% file: abstract.tex
Human activity recognition has become an attractive research area with the development of on-body wearable sensing technology. With comfortable electronic-textiles, sensors can be embedded into clothing so that it is possible to record human movement outside the laboratory for long periods. However, a long-standing issue is how to deal with \emph{motion artefacts} introduced by movement of clothing with respect to the body. Surprisingly, recent empirical findings suggest that cloth-attached sensor can actually achieve \emph{higher accuracy} of activity recognition than rigid-attached sensor, particularly when predicting from short time-windows. In this work, a probabilistic model is introduced in which this improved accuracy and resposiveness is explained by the increased statistical distance between movements recorded via fabric sensing. The predictions of the model are verified in simulated and real human motion capture experiments, where it is evident that this counterintuitive effect is closely captured.
\glsresetall

%% file: introduction.tex
Human motion analysis is critical in a wide range of research areas, from human–robot interaction \cite{anagnostis2021human} to physical rehabilitation and medical care \cite{meng2020recent}.
Recently, the development of \gls{etextiles} has made it possible to embed sensors into garments \cite{castano2014smart}. This has major advantages, such as being able to ensure the wearer’s comfort by unobtrusive sensing and allowing the capture of \emph{natural} behaviour \cite{yang_e-textiles_2019}.

However, one of the issues with clothing-embedded sensing is the additional motion of the fabric movement with respect to the body (see \fref{ma}). The prevailing view is that this motion needs to be treated as noise that should be eliminated. For this purpose, several approaches to remove or limit it have been suggested, such as \il{\item ensuring a rigid attachment between sensor and body \cite{slyper2008action}, \item supervised errors-in-variables regression \cite{michael2014eliminating}, \item unsupervised latent space learning \cite{michael2015}, \item and difference mapping distributions \cite{lorenz2022towards}}. However, recent work \cite{michael2017} suggests that fabric motion may actually \emph{assist} human motion analysis, particularly in \gls{ar}, but so far lacks a satisfactory theoretical explanation or model.

To this end, this paper proposes a probabilistic framework as a basis for understanding this phenomenon. A probabilistic model is introduced in which it can be shown that statistical distance measures such as the \gls{ks} test imply that stochastic fabric movements lead to greater discriminative ability. The predictions of the model are verified in a set of simulated and real human motion capture experiments, where it is evident that sensors loosely-attached through fabric yield greater accuracy than rigidly attached ones especially when making predictions under tight time constraints. Despite the simplicity of the model, the results show it is surprisingly accurate at capturing this phenomenon in a variety of conditions, suggesting it could be a useful tool in the design and analysis of motion capture systems using ordinary garments to enjoy their comfort and user-acceptability.

 


\begin{figure}
\begin{overpic}[width=\linewidth]{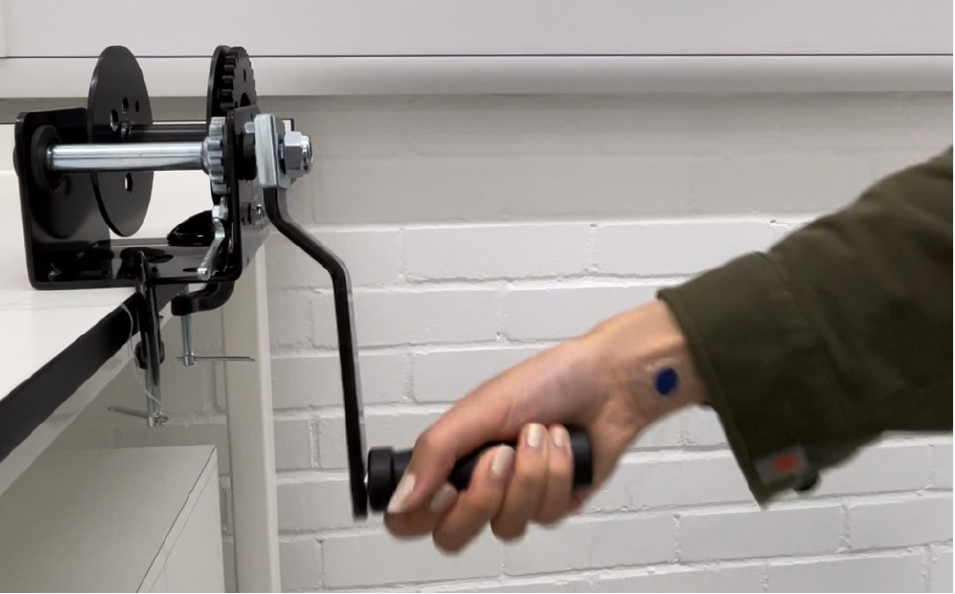}
\begin{tikzpicture}
\coordinate (ref) at (0,0);
\draw [ultra thick,dashed](1,4.05)  -- (5.5,4.1) node [near end] {\AxisRotator};
\draw [ultra thick, blue](ref)+(5.9,2.2) -- (5.4,3);
\put (58,35){\huge $\Xr$}
\draw [ultra thick, red](ref)+(7.1,1) -- (6.6,0.3);
\put (70,1){\huge $\Xf$}
\draw [color=black!60!green,decorate, thick,decoration=snake] (6,3) -- (7,3.4);
\draw [color=black!60!green,decorate, thick,decoration=snake] (7,3.6) -- (8.3,4.3);
\draw [color=black!60!green,decorate, thick,decoration=snake] (7.3,0.7) -- (8,1.1);
\draw [color=black!60!green,decorate, thick,decoration=snake] (8.1,1) -- (8.5,1.3);
\draw [color=black!60!green,decorate, thick,decoration=snake] (7.6,0.5) -- (8.5,0.8);
\draw [color=black!60!green,decorate, thick,decoration=snake] (6.3,1.4) -- (6.6,0.8);
\end{tikzpicture}
\end{overpic}~%
\caption{New sensing technologies has lead to the potential to capture human movement from clothing-embedded sensors $\Xf$ instead of relying on those rigidly attached to the body $\Xr$. However, use of ordinary garments, exposes the former to additional, unpredictable artefacts in the signal.}
\label{f:ma}
\end{figure}
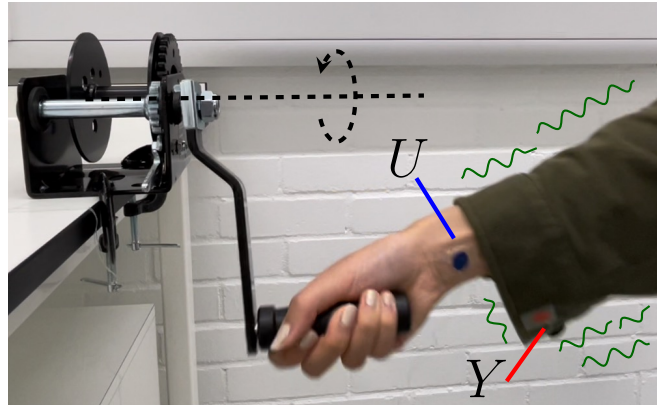

%% file: problem_definition.tex
The following introduces the proposed framework  and presents a worked example of its application to a simple movement classification task.
\subsection{Problem Definition}
\Acrlong{ar} is defined as the classification of movements into a number of discrete categories of activity (\eg walking, running, \etc) based on motion data. For simplicity, in the following, it is assumed that the latter, $\Df$, consists of $\Nd$ measurements of the absolute position of a point $\pf$ on a garment, collected over an extended duration of time, at a regular sampling frequency. In \gls{ar}, each data point also contains a label\footnote{Throughout the paper, without loss of generality, the class labels are assumed to be binary.} $\y\in\{0,1\}$ corresponding to the category of movement, so $\DfAR=\{(\bxf_1,\y_1),...,(\bxf_\Nd,\y_\Nd)\}$. This is contrasted with so-called \emph{rigid data} $\Dr$, recorded under the same conditions, except that the sensor is rigidly attached to the moving body. The goal of \gls{ar} is to train a classifier on $\DfAR$ such that, when presented with (previously unseen) movement data $\bxfd$, the corresponding class label $\yd$ can be accurately predicted.

Previous studies have provided empirical evidence that, contrary to expectations, \gls{ar} performance is \emph{improved} when using data from \emph{loose fitting garments} \cite{michael2017}. This paper tests the hypothesis that this effect is due to \emph{artefacts} introduced by the motion of the fabric, thereby simplifying the task of distinguishing between movement categories. 

\subsection{Probabilistic Model of Fabric Motion}\label{s:fabric_motion_model}
The effect of fabric motion on the data $\Df$ is subject to a high degree of uncertainty, arising from challenges in estimating the fabric's physical properties and the resultant movement complexity. To deal with this, it is proposed to model the data generation process through stochastic methods.

Specifically, the position $\bxf$ of the point $\pf$ on the fabric at any given time is modelled as the stochastic process consisting of the corresponding position $\bxr$ of a point $\pr$ on the rigid body plus a random offset $\bdelta$ introduced by the fabric motion.
In the univariate case, this can be written as
\begin{equation}
\Xf = \Xr + \Xd
	\label{e:Xf}
\end{equation}
where $\xf(t)\sim\Xf$, $\xr(t)\sim\Xr$ and $\delta(t)\sim\Xd$.

The key to determining whether the fabric movement is beneficial to \gls{ar} is to understand the effect of $\Xd$ on the statistical distance between movement classes. A condition for improved classification performance is a greater statistical distance in data distribution between movement classes in $\Df$ compared to $\Dr$, \ie
\begin{equation}
\Dd(\XfyO,\Xfyl) > \Dd(\XryO,\Xryl)
	\label{e:DXfXr}
\end{equation}
where $\XfyO$ is the distribution of fabric data for movement class $\y=0$, $\Xfyl$ is the same for movement class $\y=1$, $\XryO$ and $\Xryl$ are the equivalent distributions for rigid data, and $\Dd(\cdot,\cdot)$ is a suitable metric in probability space. For the latter, several choices are available (\eg \gls{kl} divergence or \gls{js} divergence). 
In the following, the \gls{ks} metric is used \cite{justel1997multivariate}
\begin{equation}
	\begin{split}
&\Dd(\XfyO,\Xfyl)=\text{sup}\left| \FXfyO-\FXfyl\right|\\
&\Dd(\XryO,\Xryl)=\text{sup}\left| \FXryO-\FXryl\right|\\
	\end{split}
	\label{e:ks_def}
\end{equation}
where $\F(\cdot)$ denotes the \gls{cdf} of a random variable. Note that, \gls{ks} is chosen since the range of possible fabric positions may differ depending on the movement, meaning that $\XfyO$ and $\Xfyl$ occupy different probability spaces, preventing measures such as \gls{kl} or \gls{js} from being computed.
In the next section, a simple worked example is presented to illustrate the effect of fabric motion on \acrlong{ar} predicted by this model.

\subsection{Example: Oscillatory Motion}\label{s:shm}
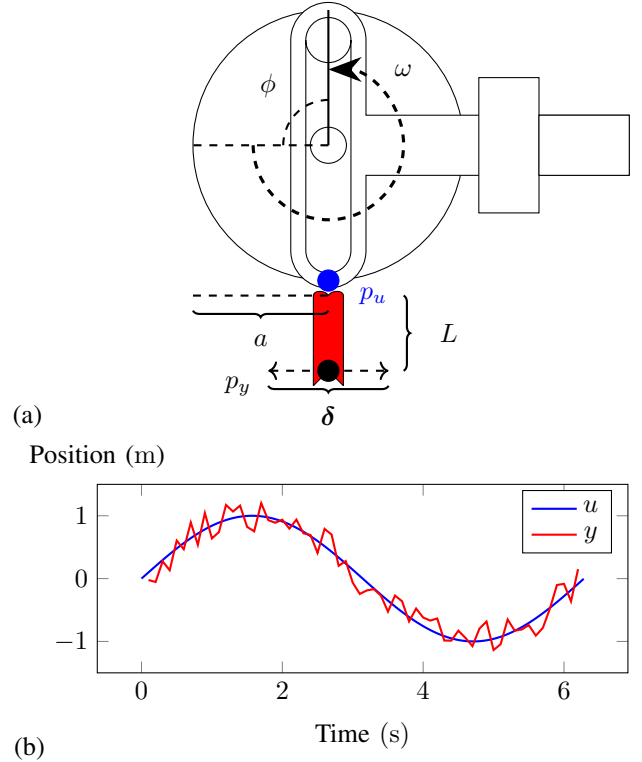
\begin{figure}[h]
\centering
\input{figure/yoke.tex}%

\input{figure/position.tex}%
\caption{\cl{\item\label{f:simulation-a} A scotch yoke mechanism with a piece of fabric attached. Sensors are affixed at $\pr$ (the tip of the sliding yoke) and $\pf$ (the tip of the fabric). \item\label{f:simulation-b} Simulated signals of the sensors from this model ($\v\L = 0.3$).}}
\label{f:simulation}%
\end{figure}

Consider the problem of movement analysis for a one dimensional, scotch yoke mechanism using data from a sensor mounted on an inextensible piece of fabric attached to the mechanism (see \fref{simulation}\ref{f:simulation-a}). Here, an \gls{ar} task might involve classifying a set movements of the rigid body (\eg those with different frequencies) from raw positional data.

In this system, the horizontal position of the rigid body (point $\pr$) over time is given by
\begin{equation}
\xr(t) =\Ay\sin(\w t+\phiy)
\label{e:xr}
\end{equation}
where $\Ay$ is the amplitude
, $\w$ is the frequency (equivalently, the angular speed of the rotary wheel of the mechanism) and $\phiy$ is the phase (equivalently, the starting angle of the wheel) and (without loss of generality) it is assumed $\Ay=1$. In this system, samples of the rigid body position follow a beta distribution
\begin{equation}
	\Xr\sim\B(\frac{1}{2},\frac{1}{2}).
	\label{e:Xr}
\end{equation}

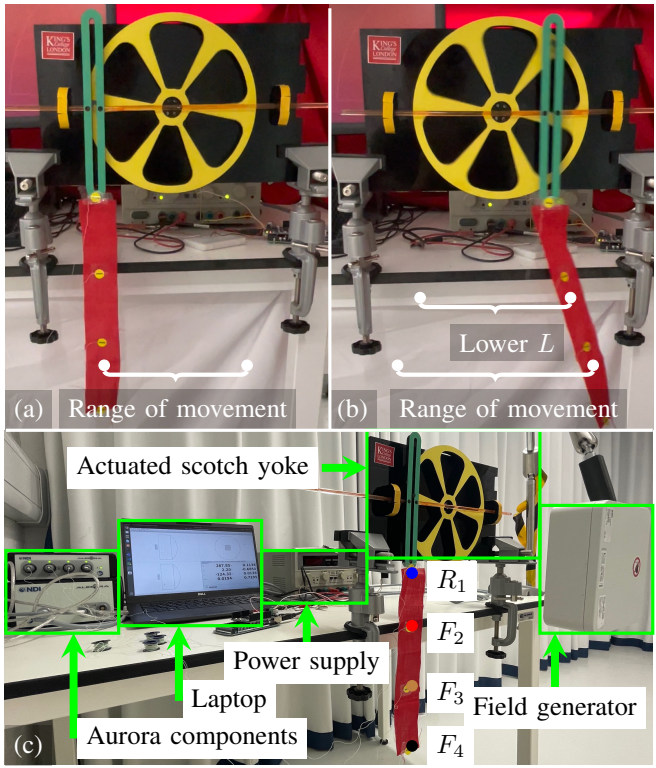
\begin{figure}[thpb]
\begin{overpic}
[width=1\linewidth]{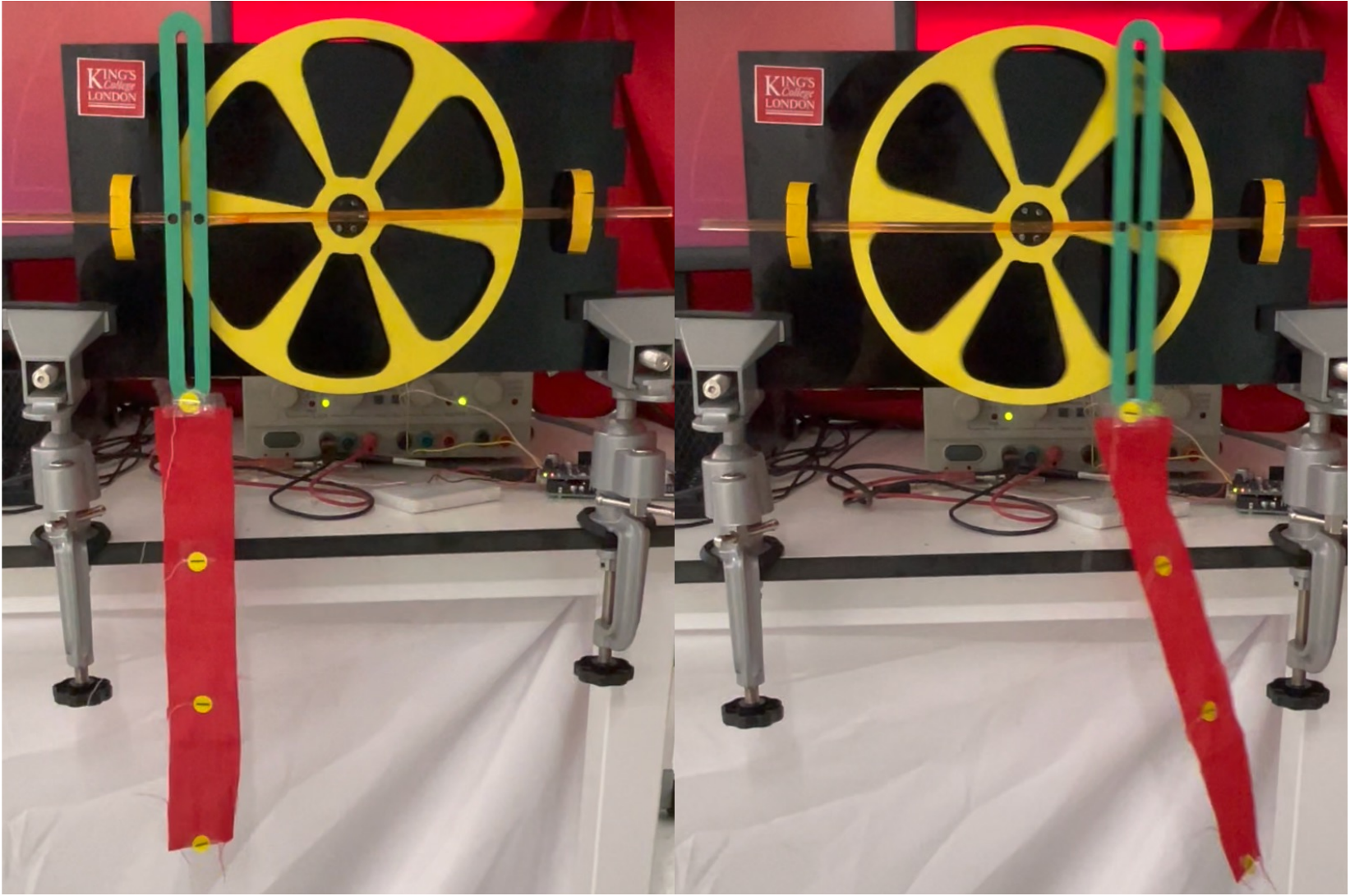}
\begin{tikzpicture}
\coordinate (ref) at (0,0);
\draw[decorate,decoration={brace,mirror,raise=8pt},ultra thick,white]  (ref)+(1.3,1.1) --  (3.2,1.1);
\node at (0.7,0.3)[right,fill=black,text=white,opacity=.25,text opacity=1]{Range of movement};
\filldraw[white] (ref)+(1.3,0.9) circle (2pt) node[anchor=west]{};
\filldraw[white] (3.2,0.9) circle (2pt) node[anchor=west]{};
\node at (0.3,0.3)[fill=black,text=white,opacity=.25,text opacity=1]{\ref{f:setup-a}};
\node at (4.6,0.3)[fill=black,text=white,opacity=.25,text opacity=1]{\ref{f:setup-b}};
\draw[decorate,decoration={brace,mirror,raise=8pt},ultra thick,white]  (ref)+(5.2,1.1) --  (7.8,1.1);
\node at (5.1,0.3)[right,fill=black,text=white,opacity=.25,text opacity=1]{Range of movement};
\filldraw[white] (ref)+(5.2,0.9) circle (2pt) node[anchor=west]{};
\filldraw[white] (7.8,0.9) circle (2pt) node[anchor=west]{};
\draw[decorate,decoration={brace,mirror,raise=8pt},ultra thick,white]  (ref)+(5.5,2) --  (7.5,2);
\node at (5.9,1.2)[right,fill=black,text=white,opacity=.25,text opacity=1]{Lower $\L$};
\filldraw[white] (5.5,1.8) circle (2pt) node[anchor=west]{};
\filldraw[white] (7.5,1.8) circle (2pt) node[anchor=west]{};
\draw[white,very thick]  (ref)+(4.3,0) -- (4.3,5.65);
\end{tikzpicture}
\end{overpic}
\input{figure/setup}
\caption{The scotch yoke mechanism with a piece of fabric attached moving at \cl{\item \label{f:setup-a} low and \item \label{f:setup-b} high frequency. \item \label{f:setup-c} Front-view of the experiment set up.} Sensors are placed \il{\item at equidistant intervals along the fabric strip ($\df_{2}$-$\df_{4}$) and \item rigidly at the attachment point ($\dr_{1}$)}.}
\label{f:setup}
\end{figure}

As the rigid body moves, the fabric will move alongside it, but undergo additional motion due to the complex fabric dynamics according to \eref{Xf}. 
In general, the nature of the fabric motion will depend on \il{\item its physical properties (\eg mass distribution, stiffness, fibre structure, length, width, orientation) and \item the movement pattern of the rigid body (\eg amplitude, frequency)}. Noting that, here, the fabric is inextensible, the maximum possible displacement $\xd$ of the point $\pf$ from $\pr$ is $\L$ (without loss of generality, it is assumed that $0<\L\leq1$), suggesting that a simple choice of its distribution could be
\begin{equation}
\Xd\sim\U(-\L,\L).
\end{equation}
However, this would imply that any displacement $-\L\le\xd\le\L$ is equally likely, whereas in practice, $\xd$ tends to be greater when there is greater excitation of the fabric by the movement of the yoke, \eg for higher-frequency movements (see \fref{setup}\ref{f:setup-a}\ref{f:setup-b}). Therefore, the following assumes
\begin{equation}
\Xd\sim\U(-\v\L,\v\L)
\end{equation}
%

%
%
where
\begin{equation}
\v = 1-\exp(-\w^2).
	\label{e:z}
\end{equation}
Note that, this respects the constraint that $-\L\le\delta\le\L$ (since \eref{z} causes $0\le\v\le1$) while capturing the tendency for $\delta$ to increase at higher frequencies up to a saturation point.
According to this model, the \gls{cdf} of $\Xr$ is \cite{shin2008fundamentals} 
\begin{equation}
F(\xr)=
\begin{cases}
	\frac{\sin^{-1}(\xr)}{\pi}+\frac{1}{2} &-1\leq \xr \leq 1\\
0& \text{otherwise}\\
\end{cases}
\label{e:cdf:sine}
\end{equation}
and it can be shown 
that the \gls{cdf} of $\Xf$ is
\begin{equation}
	F(\xf)=
	\begin{cases}
		0        &            \xf<-1-\v\L\\
		F_1(\xf) & -1-\v\L\leq\xf<-1+\v\L\\
		F_2(\xf) & -1+\v\L\leq\xf\leq 1-\v\L\\
		F_3(\xf) &     1-\v\L<\xf\leq\v\L+1\\
		1        &            \xf>1+\v\L 
	\end{cases}
	\label{e:CDFXf}
\end{equation}
where %
\begin{equation}
	F_1(\xf)=\frac{\pi\vp+2\sqrt{1-\vp^2}+2\vp\sin^{-1}\vp}{4\v\L\pi},%
\end{equation}%
\begin{equation}%
\begin{split}%
	F_2(\xf)&=\frac{1}{2}+\\&\frac{\sqrt{1-\vp^2}-\sqrt{1-\vm^2}+\vp\sin^{-1}\vp-\vm\sin^{-1}\vm}{2\v\L\pi},\\%
\end{split}%
\end{equation}%
\begin{equation}%
\begin{split}%
F_3(\xf)=\frac{\frac{3 \pi \v L + \pi \xf}{2}-\vm sin^{-1}(\vm)-\sqrt{1-\vm^2}}{2\v L \pi}\\
\end{split}%
\end{equation}%
and $\v_\pm=\v\L\pm\xf$. The \gls{pdf} and \gls{cdf} of $\Xr$ and $\Xf$ are shown in \fref{pro}\ref{f:pdf} and \ref{f:cdf}, respectively. As can be seen in \fref{pro}\ref{f:pdf}, the range of fabric positions is larger than that of the rigid body, meaning they occupy different probability spaces.

\begin{figure}[thpb]
\input{figure/pdf}%
\input{figure/cdf}%
\caption{The \cl{\item\label{f:pdf} \gls{pdf}  and \item\label{f:cdf} \gls{cdf} of rigid body and fabric position ($\v\L = 1$).}}%
\label{f:pro}%
\end{figure}%
From \eref{cdf:sine} it is apparent that the \gls{cdf} of $\Xr$ is \emph{independent of $\w$}, meaning that if the \gls{ar} task is to classify movements of different frequency (\eg $\w_1$ and $\w_2$) \emph{the data distribution can not increase statistical distance to discriminate between classes}, \ie $\Dd(\Xr_{\w_1},\Xr_{\w_2})
=0$. However, from \eref{CDFXf}, it can be shown that
\begin{equation}
	\begin{split}
\Dd(\Xf_{\w_1},\Xf_{\w_2}) &= \text{sup}\left| F(\Xf_{\w_1})-F(\Xf_{\w_2})\right|\\
&=\left| F(\xf_{\w_1|\xf=1+\v_{1}L})-F(\xf_{\w_2|\xf=1+\v_{1}L})\right|\\
&=\frac{\pi (\dv L - 1) +2\sqrt{1-(\dv L- 1)^2}}{4\v_{2} L \pi}\\
&-\frac{ 2(1 -\dv L)sin^{-1}(\dv L -1 )}{4\v_{2} L \pi}
		\label{e:ks:classification}
	\end{split}
\end{equation}
where $\dv = |\v_{2} - \v_{1}|$.
As $\Dd(\Xf_{\w_1},\Xf_{\w_2})>0$ the condition \eref{DXfXr} is met, suggesting that \emph{\gls{ar} based on the motion of the fabric} will lead to \emph{higher classification performance}.
Moreover, \emph{$\Dd(\Xf_{\w_1},\Xf_{\w_2})$ increases with $\L$}. To see this, note that
\begin{equation}
\begin{split}
&\frac{d\Dd(\Xf_{\w_1},\Xf_{\w_2})}{d\L}\\
&=\frac{\pi-2(\sqrt{-\L\dv(\L\dv-2)}+sin^{-1}(1-\L\dv))}{4\pi\v_{2}\L^2}.
\end{split}
\end{equation}
As $\max\{2(\sqrt{-\L\dv(\L\dv-2)}+sin^{-1}(1-\L\dv))\}=\pi$, it is clear that $\frac{d\Dd(\Xf_{\w_1},\Xf_{\w_2})}{d\dv}>0$ for all $\L$, so \eref{ks:classification} is a monotonically increasing function with respect to $\L$. 

This suggests that \emph{the looser the fabric, the greater the statistical distance}. 
To the authors' knowledge, this is the first analytical model to capture and explain the empirical finding that data from loose clothing can lead to enhanced \gls{ar}.
An empirical study of this example is provided in the next section.

%% file: figure/yoke.tex
\begin{tikzpicture}[scale=2]
  \draw (0,0) circle (0.9);
  \filldraw[fill=white,rotate=90] (0.7,0)++(0,0.25) -- ++(-1.4,0) arc (90:270:0.25) -- ++(0.5,0) |- ++(0.4,-1.75)  |- ++(0.5,1.75) arc (-90:90:0.25);
  \draw (0,0) circle (0.12);
  \draw [rotate=90](0.7,0) circle (0.15) ++(0,0.15) -- ++(-1.4,0) arc (90:270:0.15) -- ++(1.4,0);
  \filldraw[fill=white,draw=black,rotate=90] (-0.45,-1) rectangle ++(0.9,-0.4);
  \fill[fill=white,draw=black,rotate=90] (-0.2,-2) rectangle ++(0.4,0.6);
  \filldraw[fill=red,draw=black,rounded corners=0.2,yshift=1cm] (-0.1,-2) to [out=80,in=120] (0,-2) to [out=60,in=100] (0.1,-2) -- ++(0,-0.6) -- (0,-2.5) -- (-0.1,-2.6) -- cycle;
  \node at (-0.6,-1.6){$\pf$};%
  \node [color=blue] at (0.3,-1){$\pr$};%
  \draw[decorate,decoration={brace,mirror},thick] (0.5,-1.5) -- (0.5,-1);
 \node at (0.8,-1.25){$\L$};%
 \filldraw[blue] (0,-0.9) circle (2pt) node[anchor=west]{};
 \filldraw[black] (0,-1.5) circle (2pt) node[anchor=west]{};
\draw[very thick, dashed,arrows = {-Stealth[scale=1.5]} ] (-0.5,0) arc (0.5:270:-0.5);
\node at (0.5,0.5){$\w$};%
\draw [dashed, thick,<-] (-0.4,-1.5) -- (0,-1.5);
\draw [dashed, thick,->] (0,-1.5) -- (0.4,-1.5);
\draw [decorate,
    decoration = { brace,mirror},thick] (-0.4,-1.6) --  (0.4,-1.6);
\node at (0,-1.8){$\bdelta$};%
\draw [dashed, thick] (-0.9,-1) -- (0,-1);
\draw [decorate,
    decoration = { brace,mirror},thick] (-0.9,-1.1) --  (0,-1.1);
\node at (-0.45,-1.3){$\Ay$};%
\draw [dashed, thick] (0,0) -- (-0.9,0);
\draw [thick] (0,0) -- (0,0.9);
\draw[thick,dashed] (-0.3,0) arc (0.3:-90:-0.3);
\node at (-0.4,0.4){$\phi$};%
\node at  (-2,-1.8){\ref{f:simulation-a}};
\end{tikzpicture}

%% file: figure/position.tex
\begin{tikzpicture}
\begin{axis}[
xlabel=Time $(\si{\second})$,
ylabel=Position ($\si{\meter}$),
every axis y label/.style={at={(current axis.north west)},above=1mm},
legend pos=north east,
ymax=1.5,
ymin=-1.5,
width= 1\linewidth,
height = 0.23\textwidth,
every axis plot/.append style={thick},
]
\addplot[smooth,blue,mark=none,
domain=0:2*3.14,samples=63]
{sin(deg(x))};
\addplot[red] file[skip first] {data/fabric_pos.dat};
\addlegendentry{$\xr$}
\addlegendentry{$\xf$}%
\end{axis}
\node at  (-0.9,-1){\ref{f:simulation-b}};
\end{tikzpicture}

%% file: figure/setup.tex
\begin{overpic}[width=1\linewidth]{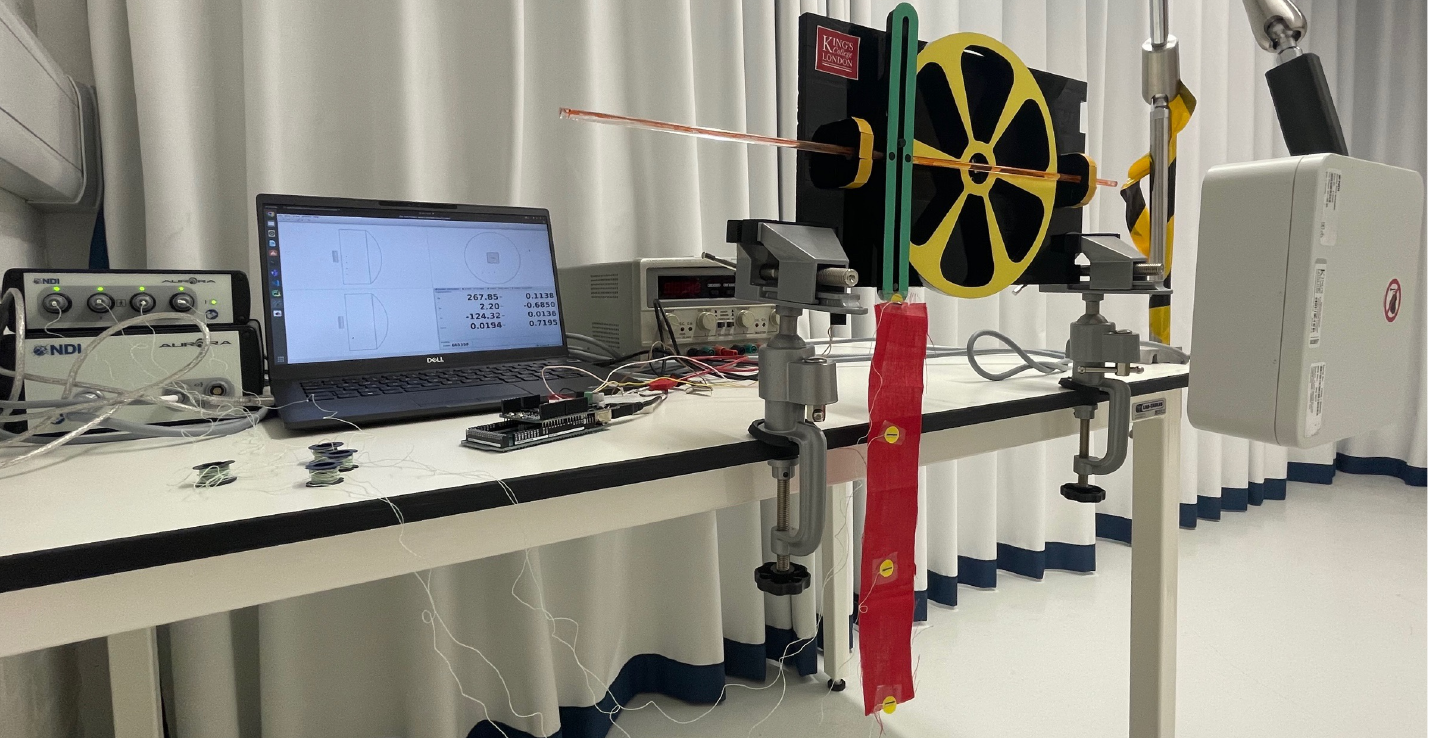}
\begin{tikzpicture}
\draw [color=green,very thick] (-4.4,-0.5) rectangle (-2.9,0.6);
\draw [color=green,very thick] (-2.9,-0.4) rectangle (-1,1); 
\draw [color=green,very thick] (-1,-0.1) rectangle (0.4,0.6); 
\draw [color=green,very thick] (0.4,0.5) rectangle (2.7,2.2); 
\draw [color=green,very thick] (2.7,-0.5) rectangle (4.2,1.2); 
\draw[stealth-,line width = 3pt,green] (-3.5,-0.5) -- (-3.5,-1.9) node[right,black,fill=white]{Aurora components};
\draw[stealth-,line width = 3pt,green] (-2.1,-0.4) -- (-2.1,-1.4) node[right,black,fill=white]{Laptop};
\draw[stealth-,line width = 3pt,green] (-0.4,-0.1) -- (-0.4,-0.9) node[black,fill=white]{Power supply};
\draw[stealth-,line width = 3pt,green] (0.4,1.7) -- (-0.2,1.7) node[left,black,fill=white]{Actuated scotch yoke};
\draw[stealth-,line width = 3pt,green] (2.9,-0.5) -- (2.9,-1.5) node[black,fill=white]{Field generator};
\filldraw[blue] (1,0.3) circle (2pt) node[anchor=west]{};
\filldraw[red] (1,-0.4) circle (2pt) node[anchor=west]{};
\filldraw[brown] (1,-1.2) circle (2pt) node[anchor=west]{};
\filldraw[black] (1,-2) circle (2pt) node[anchor=west]{};
\node at (1.2,0.2)[right,black,fill=white]{$\dr_{1}$};
\node at (1.2,-0.5)[right,black,fill=white]{$\df_{2}$};
\node at (1.2,-1.3)[right,black,fill=white]{$\df_{3}$};
\node at (1.2,-2)[right,black,fill=white]{$\df_{4}$};
\node at (-4.1,-2)[fill=black,text=white,opacity=.25,text opacity=1]{\ref{f:setup-c}};
\draw[white,very thick]  (-4.4,2.2) -- (4.2,2.2);
\end{tikzpicture}
\end{overpic}

%% file: figure/pdf.tex
\begin{tikzpicture}
    \begin{axis}[
axis y line=middle,%
axis x line=middle,%
xmin = -2,
xmax = 2,
ymin = 0,
ymax = 1,
ytick={0,0.5,1},
xtick={-2,-1,0,1,2},
width= 0.95\linewidth,%
height = 0.5\linewidth,%
ylabel=$f$,
scale only axis, 
        > = Stealth
    ]
        \addplot[
            blue,
           very thick,
            domain = -1:1,
            samples = 1000
        ]
            {1/(pi*sqrt(1-x*x))};
        \addplot[
            blue,
            very thick,
            dashed,
            mark = none
        ]
            coordinates {(-1,0) (-1,1)};
        \addplot[
            blue,
            very thick,
            dashed,
            mark = none
        ]
            coordinates {(1,0) (1,1)};
    \addplot[red,very thick] file[skip first] {data/mass_pdf.dat};%
    \end{axis}
    \node at (0,-0.9){\ref{f:pdf}};%
\end{tikzpicture}

%% file: figure/cdf.tex
\begin{tikzpicture}
\begin{axis}[%
axis y line=middle,%
axis x line=middle,%
xlabel=Position,
ylabel=$F$,%
legend pos=north west,
ymax=1,%
ymin=0,%
xmax=2,%
xmin=-2,%
width= 0.95\linewidth,%
height = 0.5\linewidth,%
scale only axis,%
ytick={0,0.5,1},
xtick={-2,-1,0,1,2},
every axis plot/.append style={thick},
legend style={font=\small},
]%
\addplot[blue,very thick] file[skip first] {data/mass_cdf.dat};%
\addplot[red,very thick] file[skip first] {data/fabric_cdf.dat};%
\addlegendentry{$\Xr$}%
\addlegendentry{$\Xf$}%
\end{axis}%
\node at (0,-0.9){\ref{f:cdf}};%
\end{tikzpicture}

%% file: material_and_methods.tex
As noted in \sref{fabric_motion_model}, the extent to which fabric motion helps \gls{ar} in practice will depend on both the \emph{complexity of the movement} and the \emph{physical properties of the fabric}. In this section, \ref{N_case_studies} empirical case studies are presented to test the model's predictions when using a well-established statistical machine learning approach for \gls{ar} with fabric-induced motion data. The cases considered are \il{\item a numerical simulation of example described in \sref{shm}\item\label{v:scotch}its physical realisation and \item\label{v:human} a real human \gls{ar} task}. \footnote{Data and source code for these experiments is available online at \url{http://[url_available _on_acceptance]}.}

\newcounter{casestudy}
\stepcounter{casestudy}%
\subsection{Case Study \thecasestudy: Simple Harmonic Motion}\label{s:sim}
%
%
This evaluation aims to verify the predictions of the proposed model using a numerical simulation. %

\subsubsection{Materials and Methods}\label{s:scotch_yoke:mnm}
Data is collected from a numerical simulation of the system shown in \fref{simulation}\ref{f:simulation-a} implemented in MATLAB R2019b (MathWorks, USA) consisting of 
trajectories of length $\Tk=\TkscotchYokeSimulation$\,\si{\second} 
generated at a sampling rate of $\fk=\SI{\fscotchYokeSimulation}{\Hz}$ using \eref{Xf} from random starting angles $\phiy \sim\U(-\pi,\pi)$ with $\Ay=\SI{\AscotchYokeSimulation}{\metre}$. 
Each data set contains $\Nd=\NdscotchYokeSimulation$ trajectories with $\NdscotchYokeSimulationLo$ trajectories from the yoke running at low-frequency (\ie $\w_{1}=1\,\si{\radian\per\second}$
) and $\NdscotchYokeSimulationHi$ at high-frequency ($\w_{2}=2\,\si{\radian\per\second}$
). The same procedure is used to collect fabric movement data $\Df$ for lengths $\L\in\{\frac{1}{3},\frac{2}{3},1\}\si{\metre}$ (\ie $\df_{1}$, $\df_{2}$, $\df_{3}$) and rigid body data $\Dr$ (\ie $\dr_{1}$). 
The data is split into equal-sized training and test sets and used to train a \gls{svm} classifier (Libsvm toolbox \cite{chang_libsvm_2011}) to perform \gls{ar} with Gaussian \glspl{rbf} as the kernel function. The \gls{svm} is trained to predict the mapping
\begin{equation}
\bphi_\nd\mapsto\y_\nd
\end{equation}
in an online fashion, where $\bphi_\nd$ is a fragment of the $\nd$th trajectory and $\y_\nd\in\{0,1\}$ is the corresponding class label ($c=0$ for $\w_1$, $c=1$ for $\w_2$).
%
Specifically, following \cite{michael2017}, each trajectory is segmented into overlapping windows of size $\Nw-1$ (where $\Nw < \Nk $ and $\Nk=\Tk/\fk$), \ie
\begin{equation}
\begin{split}
\bPhi:&=\left(\bphi_1,\bphi_2,...\bphi_\Nd\right)\\
&=\left((\xf_{1},...,\xf_{\Nw})^\T,(\xf_{2},...,\xf_{\Nw+1})^\T,(\xf_{3},...,\xf_{\Nw+2})^\T,...\right).
\end{split}
	\label{e:bphinsy}
\end{equation}%
The procedure is repeated $\TrialsscotchYokeSimulation$ times for each condition and the classification accuracy computed. 
\subsubsection{Results}\label{s:scotch_yoke:results}
\fref{result}\ref{f:result-a} shows the overall accuracy of \gls{ar} using the \gls{svm} classifier with different window sizes. As can be seen, the accuracy is higher when using fabric data at small window sizes. Moreover, the accuracy is higher when $\L$ is greater, in line with the prediction of the model. As the window size increases (\ie the classifier is given more of the trajectory history) the overall accuracy increases and the difference between fabric and rigid data gradually disappears.

\stepcounter{casestudy}%
\subsection{Case Study \thecasestudy: Scotch yoke}\label{s:scotch_yoke}
\label{s:Experiment}
\label{s:data_acquisition}
This evaluation aims to validate the proposed framework in a \emph{physical system}. 
\subsubsection{Materials and Methods}\label{s:scotch_yoke:mnm}
To ensure accurate and repeatable data collection, the experiment reported here uses an instrumented \emph{scotch yoke} as a data acquisition device. 
%
%
The experimental setup is shown in \fref{setup}\ref{f:setup-c}. The actuated scotch yoke consists of a sliding yoke, with rigid rods affixed either side and a rotating disk of diameter $\SI{20}{\cm}$ mounted on two bearing blocks and driven by a DC motor with encoder ($30:1, 37D$ gear-motor, Pololu Corporation, USA) at the fulcrum. The motion of the disk and yoke are coupled via a sliding pin, ensuring a pure sinusoidal movement of the yoke. Affixed to the latter, $\SI{10}{\cm}$ away from the fulcrum, is a \SI{30}{\cm}$\times$\SI{5}{\cm} 
strip of woven cotton fabric 
upon which are mounted three sensors (NDI Aurora Magnetic Tracking device, NDI, Canada) that synchronously record the horizontal position  
at $\SI{40}{\Hz}$, at an accuracy of approximately $0.09\si{\mm}$. The latter are attached along the length of the fabric \il{\item $\SI{20}{\cm}$ ($\df_{2}$), \item $\SI{30}{\cm}$ ($\df_{3}$) and \item $\SI{40}{\cm}$ from the fulcrum ($\df_{4}$) (\ie at the tip of the fabric)}. A further sensor ($\dr_{1}$) is rigidly attached to the yoke at the fabric attachment point (see \fref{setup}\ref{f:setup-c}). The error in the yoke movement against the sinusiodal reference is 
$\pm 0.05\pi\si{\radian\per\second}$.

With this set up, data is collected from the device driven at the desired speed for the experimental condition (see below).\footnote{\label{fn:missing-data} Note that, at high speeds NDI device occasionally loses track of its sensors resulting in missing data (\ie gaps) within trajectories. These are filled using piecewise cubic spline interpolation.}\edit{.}{}{mh} Specifically, the following reports the effect of varying \il{\item the window size $0.025\le\Nw/\fk\le2.5\,\si{\second}$ where $\w_{1}=1.05\pi\,\si{\radian\per\second}$ and $\w_{2}=1.48\pi\,\si{\radian\per\second}$, and \item the difference in frequencies (\ie $|\w_{2}-\w_{1}|$) where $\Nw/\fk=\NwfkScotchYokeExperiment\,\si{\second}$} and $\w_{1}=1.05\pi\,\si{\radian\per\second}$
. For this, $\NdscotchYokeExperiment$ sample trajectories of length $\Tk=$\SI{5}{\second} at each speed are recorded. 
This data is segmented for online learning through a similar procedure as described in \sref{scotch_yoke}, randomly split into equal-sized training and test sets, and used to train a \gls{svm} classifier to perform \gls{ar}. All data are standardised using the $z$-score. 
The performance of the classifier is assessed by computing its accuracy and the \gls{ks} test statistic via \eref{ks_def}. This process is repeated for $\TrialsscotchYokeExperiment$ trials of every experimental condition tested. 

\subsubsection{Results}\label{s:scotch_yoke_experiment_results}
\begin{figure}%
\input{figure/svm_sim}%
\input{figure/svm_real}%
\input{figure/vary_frequency}%
\input{figure/vary_frequency_ks}%
\caption{Accuracy in recognising movements of different frequency in the \cl{\item\label{f:result-a} simulated and \item\label{f:result-b} physical scotch yoke when varying window size, and accuracy\item\label{f:result-c} and \item \label{f:result-d} \gls{ks} test statistic $\Dd$ for different pairs of frequencies with window size $0.025s$. Reported are the mean$\pm$ s.d. over $\TrialsscotchYokeSimulation$ trials.}}
\label{f:result}%
\end{figure}
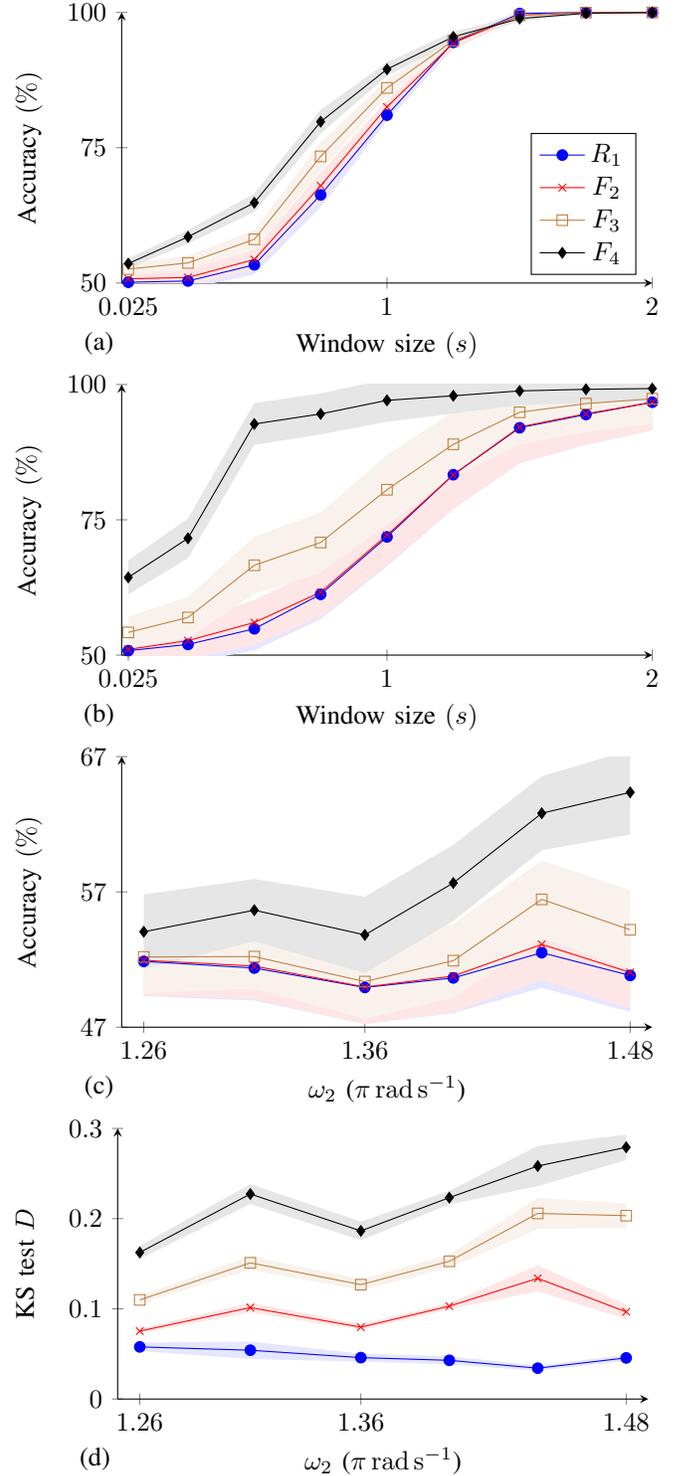

\fref{result}\ref{f:result-b} shows the accuracy of \gls{ar} using the \gls{svm} classifier for different window sizes. As can be seen, a similar, but more pronounced trend is seen here as predicted by the simulation in \sref{scotch_yoke} (\qv): the accuracy is higher for fabric-attached sensors at small window sizes (and higher for sensors at greater $\L$). As the window size increases, the accuracy converges toward the same value, regardless of the sensor used.

\fref{result}\ref{f:result-c} and \fref{result}\ref{f:result-d} show the classification accuracy and \gls{ks} test statistic D for discriminating between $\w_{1}=1.05\pi\,\si{\radian\per\second}$ and the different $\w_{2}$ when $\Nw=\SI{0.025}{\second}$ respectively.
As can be seen, for the fabric-mounted sensors, the statistical distance $\Dd$ is larger. Moreover, the larger the difference between movement frequencies, the greater the statistical distance, and therefore the better the performance of \gls{ar} (\gls{pcc} between $\Dd$ and accuracy is $ 0.85$, indicating a strong positive relationship). For the rigidly-attached sensor, there is no obvious increase in $\Dd$, nor \gls{ar} performance. This is consistent with the prediction of \eref{DXfXr}. 

\stepcounter{casestudy}%
\subsection{Case Study \thecasestudy: Human \acrlong{ar}}\label{s:human_ar}
In this section, the predictions of the proposed framework are evaluated in a human motion recognition task.\footnote{The experiments reported here were conducted with the ethical approval of King's College London, UK: MRPP-21/22-33739.}
\subsubsection{Material and methods}
The \gls{ar} task chosen for this experiment is that of recognition of constrained periodic movements (such as operating a crank, winch, or ratchet system) from loose, sensorised clothing. The experimental set up (see \fref{hand}) consists of a \emph{hand winch} with a \SI{15}{\cm} rotating crank handle that the experimental subject must operate at pre-specified speeds. To control the speed, a display screen is used to show the desired and actual position of the wrist in real time (as red and blue points, respectively), and the participant is asked to move such that, as far as possible, these coincide throughout the movement. The average error in the hand movement against the target is approximately $0.1\pi\si{\radian\per\second}$. 
During data collection, the participant wears a loose shirt ($96\%$ woven cotton and $4\%$ spandex) with one Aurora sensor attached to the sleeve (with a maximum possible displacement from the wrist of $\pm \SI{10}{\cm}$)  and a second affixed to the wrist to act as a baseline. With this set up, the subject is directed to operate the winch at \emph{low} ($\w_{1} = 1.25\pi\si{\radian\per\second}$) and \emph{high} frequency ($\w_{2} = 2.5\pi\si{\radian\per\second}$). To accustomise the participant to the set up, they are directed to perform the target movement for $15$ trials at each frequency prior to the experiment.
After that, \numberstringnum{\NdhandExperiment} sample trajectories 
of length $\Tk = \SI{8}{\second}$ at high frequency are recorded followed by the same number of samples at low frequency. The participant is given $20\si{\second}$ rest between each sample. The first $\SI{3}{\second}$ of each sample trajectory is discarded to allow for the time it takes the participant to adjust to follow the target accurately.The remaining data are used for \gls{ar} using the method described in \sref{scotch_yoke}.
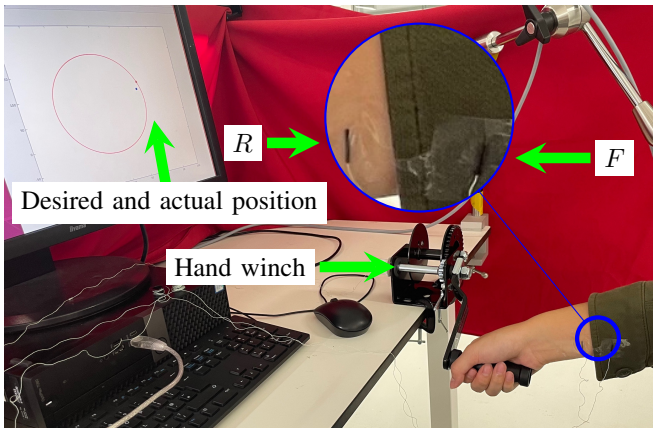
\begin{figure}[b]
\begin{tikzpicture}[spy using outlines={blue,circle,magnification=4.5, size=2.5cm, connect spies,every spy on node/.append style={ultra thick}}]
\node(1)[anchor=south west,inner sep=0,xshift=0cm,yshift=-1.7cm](image) {\includegraphics[width=\linewidth]{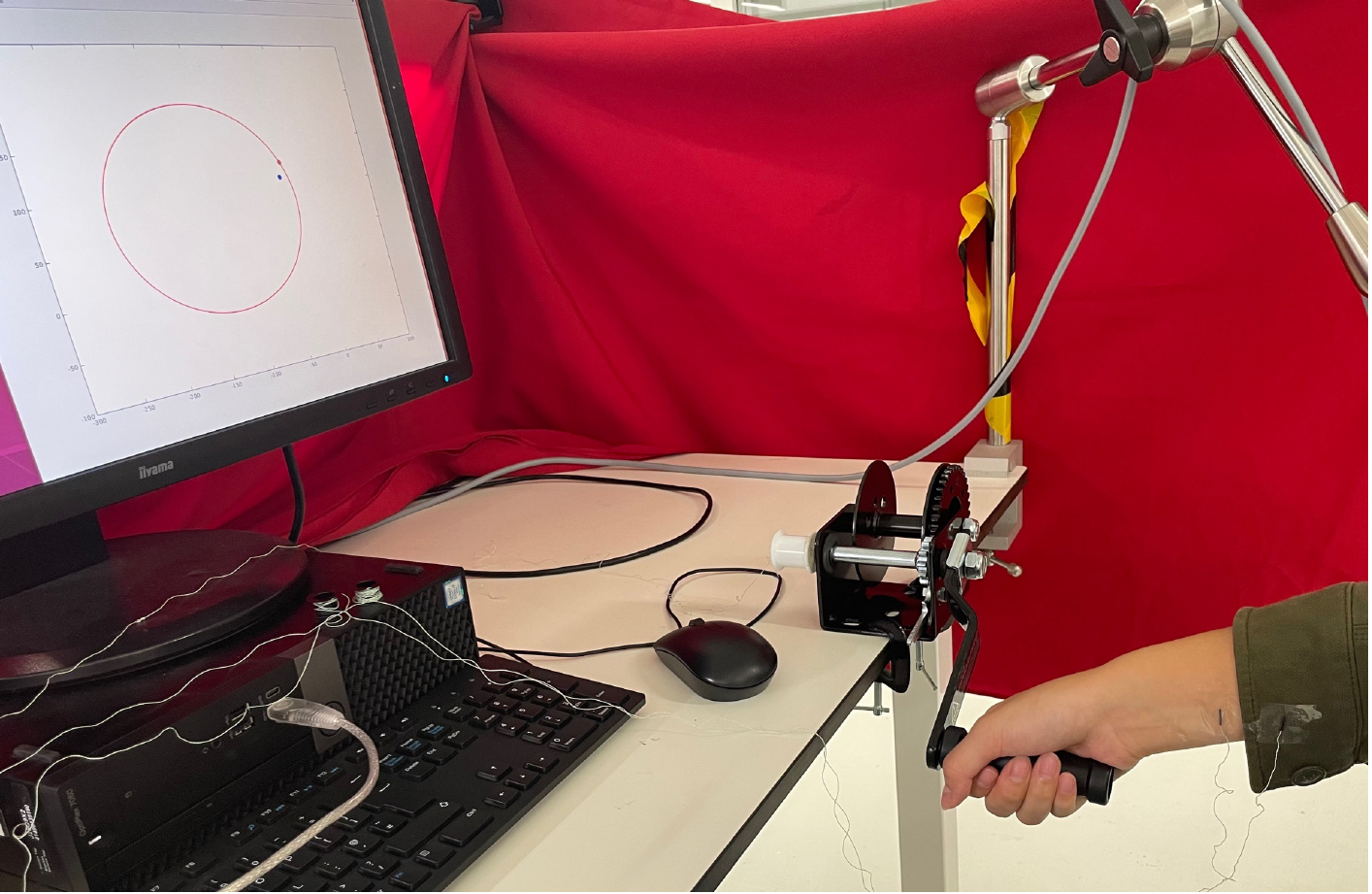}};
\spy on (7.9,-0.5) in node at ([xshift=1.2cm,yshift=-1.5cm]image.north);
\draw[stealth-,line width = 4pt,green] (4.2,2.1) -- (3.5,2.1) node[left,black,fill=white]{$\dr$};
\draw[stealth-,line width = 4pt,green] (6.8,1.9) -- (7.8,1.9) node[right,black,fill=white]{$\df$};
\draw[stealth-,line width = 4pt,green] (5.2,0.45) -- (4.2,0.45) node[left,black,fill=white]{Hand winch};
\draw[stealth-,line width = 4pt,green] (2,2.4) -- (2.2,1.3) node[black,fill=white]{Desired and actual position};
\end{tikzpicture}
\caption{Experimental set up for the crank task. The participant operates a hand winch at pre-specified speeds while their motion is captured with a wrist-attached $\dr$ and garment-mounted sensor $\df$.}
    \label{f:hand}
\end{figure}
\subsubsection{Results}
Using the \gls{svm} classifier with window size $0.025s$, the accuracy is $73.8\%\pm0.9$ ($\Dd=0.13\pm0.04$) for the wrist-attached sensor compared to $76.5\%\pm0.9$ ($\Dd=0.27\pm0.06$) for the sleeve-attached one. While the average increase in accuracy is modest in this experiment (potentially due to $\L$ being relatively small in this task), the low variance suggests that the effect is robustly in line with the predictions of the proposed model. 
\manuallabel{N_case_studies}{\numberstringnum{\thecasestudy}}

%% file: figure/svm_sim.tex
\usepgfplotslibrary{fillbetween}
\begin{tikzpicture}
  \begin{axis}[
        xlabel=Window size $(s)$,
        ylabel=Accuracy $(\%)$,
        xmin=0, xmax=2,
        ymin=50, ymax=100,
        xtick={0.025,1,2},
        ytick={50,75,100},
        legend pos=south east,
         xticklabel style=
        {anchor=near xticklabel,
         /pgf/number format/precision=3,
         /pgf/number format/fixed},
width= 1\linewidth,%
height = 0.6\linewidth,%
  axis lines = left]
\addplot [mark=*,blue]table[x=x,y=y] {l1.dat};
\addplot [mark=x,red]table[x=x,y=y] {l2.dat};
\addplot [mark=square,brown]table[x=x,y=y] {l3.dat};
\addplot [mark=diamond*,black]table[x=x,y=y] {l4.dat};
\addlegendentry{$\dr_{1}$};
\addlegendentry{$\df_{2}$};
\addlegendentry{$\df_{3}$};
\addlegendentry{$\df_{4}$};
\addplot [name path=upper,draw=none] table[x=x,y expr=\thisrow{y}+\thisrow{err}] {l1.dat};
\addplot [name path=lower,draw=none] table[x=x,y expr=\thisrow{y}-\thisrow{err}] {l1.dat};
\addplot [fill=blue!10] fill between[of=upper and lower];
\addplot [name path=upper,draw=none] table[x=x,y expr=\thisrow{y}+\thisrow{err}] {l2.dat};
\addplot [name path=lower,draw=none] table[x=x,y expr=\thisrow{y}-\thisrow{err}] {l2.dat};
\addplot [fill=red!10] fill between[of=upper and lower];
\addplot [name path=upper,draw=none] table[x=x,y expr=\thisrow{y}+\thisrow{err}] {l3.dat};
\addplot [name path=lower,draw=none] table[x=x,y expr=\thisrow{y}-\thisrow{err}] {l3.dat};
\addplot [fill=brown!10] fill between[of=upper and lower];
\addplot [name path=upper,draw=none] table[x=x,y expr=\thisrow{y}+\thisrow{err}] {l4.dat};
\addplot [name path=lower,draw=none] table[x=x,y expr=\thisrow{y}-\thisrow{err}] {l4.dat};
\addplot [fill=black!10] fill between[of=upper and lower];
\end{axis}
\node at (-0.3,-0.8){\ref{f:result-a}};%
    \end{tikzpicture}

%% file: figure/svm_real.tex
\usepgfplotslibrary{fillbetween}
\begin{tikzpicture}
    \begin{axis}[
        xlabel=Window size $(s)$,
        ylabel=Accuracy $(\%)$,
        xmin=0, xmax=2,
        ymin=50, ymax=100,
        xtick={0.025,1,2},
        ytick={50,75,100},
        legend style={legend columns=2},
        legend pos=south east,
         xticklabel style=
        {anchor=near xticklabel,
         /pgf/number format/precision=3,
         /pgf/number format/fixed},
width= 1\linewidth,%
height = 0.6\linewidth,%
  axis lines = left]
\addplot [mark=*,blue]table[x=x,y=y] {r1.dat};
\addplot [mark=x,red]table[x=x,y=y] {f2.dat};
\addplot [mark=square,brown]table[x=x,y=y] {f3.dat};
\addplot [mark=diamond*,black]table[x=x,y=y] {f4.dat};
\addplot [name path=upper,draw=none] table[x=x,y expr=\thisrow{y}+\thisrow{err}] {r1.dat};
\addplot [name path=lower,draw=none] table[x=x,y expr=\thisrow{y}-\thisrow{err}] {r1.dat};
\addplot [fill=blue!10] fill between[of=upper and lower];

\addplot [name path=upper,draw=none] table[x=x,y expr=\thisrow{y}+\thisrow{err}] {f2.dat};
\addplot [name path=lower,draw=none] table[x=x,y expr=\thisrow{y}-\thisrow{err}] {f2.dat};
\addplot [fill=red!10] fill between[of=upper and lower];

\addplot [name path=upper,draw=none] table[x=x,y expr=\thisrow{y}+\thisrow{err}] {f3.dat};
\addplot [name path=lower,draw=none] table[x=x,y expr=\thisrow{y}-\thisrow{err}] {f3.dat};
\addplot [fill=brown!10] fill between[of=upper and lower];

\addplot [name path=upper,draw=none] table[x=x,y expr=\thisrow{y}+\thisrow{err}] {f4.dat};
\addplot [name path=lower,draw=none] table[x=x,y expr=\thisrow{y}-\thisrow{err}] {f4.dat};
\addplot [fill=black!10] fill between[of=upper and lower];
\end{axis}
\node at (-0.3,-0.8){\ref{f:result-b}};%
\end{tikzpicture}

%% file: figure/vary_frequency.tex
\usepgfplotslibrary{fillbetween}
\begin{tikzpicture}
    \begin{axis}[
 xlabel=$\w_{2}$ ($\pi\,\si{\radian\per\second}$),
        ylabel=Accuracy $(\%)$,
        xmin=1.25, xmax=1.49,
        ymin=47, ymax=67,
xtick={1.26,1.36,1.48},
        ytick={47,57,67},
        legend style={legend columns=2},
        legend pos=south east,
         xticklabel style=
        {anchor=near xticklabel,
         /pgf/number format/precision=3,
         /pgf/number format/fixed},
width= 1\linewidth,%
height = 0.6\linewidth,%
  axis lines = left]
\addplot [mark=*,blue]table[x=x,y=y] {s1.dat};
\addplot[name path=upper,draw=none] table[x=x,y expr=\thisrow{y}+\thisrow{err}] {s1.dat};
\addplot[name path=lower,draw=none] table[x=x,y expr=\thisrow{y}-\thisrow{err}] {s1.dat};
\addplot[fill=blue!10] fill between[of=upper and lower];
\addplot [mark=x,red]table[x=x,y=y] {s2.dat};
\addplot[name path=upper,draw=none] table[x=x,y expr=\thisrow{y}+\thisrow{err}] {s2.dat};
\addplot[name path=lower,draw=none] table[x=x,y expr=\thisrow{y}-\thisrow{err}] {s2.dat};
\addplot[fill=red!10] fill between[of=upper and lower];
 
\addplot [mark=square,brown]table[x=x,y=y] {s3.dat};
\addplot[name path=upper,draw=none] table[x=x,y expr=\thisrow{y}+\thisrow{err}] {s3.dat};
\addplot[name path=lower,draw=none] table[x=x,y expr=\thisrow{y}-\thisrow{err}] {s3.dat};
\addplot[fill=brown!10] fill between[of=upper and lower];
 
\addplot [mark=diamond*,black] table[x=x,y=y] {s4.dat};
\addplot[name path=upper,draw=none] table[x=x,y expr=\thisrow{y}+\thisrow{err}] {s4.dat};
\addplot[name path=lower,draw=none] table[x=x,y expr=\thisrow{y}-\thisrow{err}] {s4.dat};
\addplot[fill=black!10] fill between[of=upper and lower];
    \end{axis}
    \node at (-0.3,-0.8){\ref{f:result-c}};%
    \end{tikzpicture}

%% file: figure/vary_frequency_ks.tex
\begin{tikzpicture}
    \begin{axis}[
        xlabel=$\w_{2}$ ($\pi\,\si{\radian\per\second}$),
        ylabel=\gls{ks} test $\Dd$,
        xmin=1.25, xmax=1.49,
        ymin=0, ymax=0.3,
        xtick={1.26,1.36,1.48},
        ytick={0,0.1,0.2,0.3},
   legend pos=north west,
    legend style={legend columns=2},
         axis lines = left,
width= 1\linewidth,%
height = 0.6\linewidth]
\addplot [mark=*,blue]table[x=x,y=y] {ks1.dat};
\addplot [mark=x,red]table[x=x,y=y] {ks2.dat};
\addplot [mark=square,brown]table[x=x,y=y] {ks3.dat};
\addplot [mark=diamond*,black]table[x=x,y=y] {ks4.dat};
\addplot [name path=upper,draw=none] table[x=x,y expr=\thisrow{y}+\thisrow{err}] {ks1.dat};
\addplot [name path=lower,draw=none] table[x=x,y expr=\thisrow{y}-\thisrow{err}] {ks1.dat};
\addplot [fill=blue!10] fill between[of=upper and lower];

\addplot [name path=upper,draw=none] table[x=x,y expr=\thisrow{y}+\thisrow{err}] {ks2.dat};
\addplot [name path=lower,draw=none] table[x=x,y expr=\thisrow{y}-\thisrow{err}] {ks2.dat};
\addplot [fill=red!10] fill between[of=upper and lower];
 
\addplot [name path=upper,draw=none] table[x=x,y expr=\thisrow{y}+\thisrow{err}] {ks3.dat};
\addplot [name path=lower,draw=none] table[x=x,y expr=\thisrow{y}-\thisrow{err}] {ks3.dat};
\addplot [fill=brown!10] fill between[of=upper and lower];
 
\addplot [name path=upper,draw=none] table[x=x,y expr=\thisrow{y}+\thisrow{err}] {ks4.dat};
\addplot [name path=lower,draw=none] table[x=x,y expr=\thisrow{y}-\thisrow{err}] {ks4.dat};
\addplot [fill=black!10] fill between[of=upper and lower];
    \end{axis}
    \node at (-0.3,-0.8){\ref{f:result-d}};%
    \end{tikzpicture}

%% file: discussion.tex
This work presents framework with which to understand the effect of textile motion on \gls{ar}, including how it may enhance performance compared to use of rigidly-attached sensors. By taking a statistical modelling approach, it is seen that that stochasticity in the fabric motion can \emph{amplify} the statistical distance between movement signals, \emph{enhancing \gls{ar} performance}. The predictions of this model have been verified through numerical and physical evaluations, including human motion capture for simple oscilliatory movements with the findings \il{\item \gls{ar} improves as the fabric becomes looser ($\L$ increases) and \item the discrepancy with rigid sensor use is most pronounced at small window sizes}. The latter suggests that used of fabric-mounted sensors may enable faster, more responsive predictions in the context of \emph{online} \gls{ar} and \emph{artifact} introduced by the motion of the fabric make contribute to \gls{ar}. More broadly, the fact that these effects can be analytically modelled with a relatively simple model opens up the possibility of enhanced design and analysis of motion capture systems that use ordinary garments, thereby enjoying a high level of comfort and user-acceptance.
%
In turn, this could have implications for many applications in robotics and automation, such as analysing workers' behaviour in manufacturing lines \cite{koskimaki2009activity,forkan2019industrial} and human–robot interaction (\eg for control of exoskeletons \cite{mai2021human} or prostheses \cite{pitou2018embroidered}, \etc) Future work will focus on the study of more complex human movement tasks, such as multi-joint upper limb movement or gait analysis.